\documentclass[10pt,twocolumn,letterpaper]{article}

\usepackage{iccv}
\usepackage{times}
\usepackage{epsfig}
\usepackage{graphicx}
\usepackage{amsmath}
\usepackage{amssymb}
\usepackage{algorithm}
\usepackage{algorithmic}
\usepackage{multirow}
\usepackage{makecell}
\usepackage{colortbl}
\usepackage{tabularx}

\newcommand\clearrow{\global\let\rowmac\relax}
\clearrow

\usepackage[breaklinks=true,bookmarks=false]{hyperref}
\newcommand{\by}[1]{\textcolor{black}{#1}}

\newcommand{\bp}[1]{\textcolor{black}{#1}}
\iccvfinalcopy 

\def\iccvPaperID{****} 

\ificcvfinal\fi

\begin{document}

\title{PSViT: Better Vision Transformer via Token Pooling and Attention Sharing}

\author{Boyu Chen\textsuperscript{1}\thanks{Equal contribution}, Peixia Li\textsuperscript{1}\footnotemark[1],  Baopu Li\textsuperscript{2}\footnotemark[1], Chuming Li\textsuperscript{3}, 
 Lei Bai\textsuperscript{1}, Chen Lin\textsuperscript{4}, \\ Ming Sun\textsuperscript{3}, Junjie Yan\textsuperscript{3}, Wanli Ouyang\textsuperscript{1} \\
\textsuperscript{1} The University of Sydney,
\textsuperscript{2} BAIDU USA LLC,  \\
\textsuperscript{3} SenseTime Group Limited,  
\textsuperscript{4} University of Oxford  \\
}


\maketitle
\ificcvfinal\thispagestyle{empty}\fi

\begin{abstract}
	In this paper, we observe two levels of redundancies when applying vision transformers (ViT) for image recognition.
	First, fixing the number of tokens through the whole network produces redundant features at the spatial level. 
	Second, 
	the attention maps among different transformer layers are redundant. Based on the observations above,
	we propose a PSViT: a ViT with token \textbf{P}ooling and attention \textbf{S}haring 
	to reduce the redundancy, effectively enhancing the feature representation ability, and achieving a better speed-accuracy trade-off.
	Specifically, in our PSViT, \textbf{token pooling} can be defined as the operation that decreases the number of tokens at the spatial level. 
	Besides, \textbf{attention sharing} will be built between the neighboring transformer layers for reusing the attention maps having a strong correlation among adjacent layers.  
	Then, a compact set of the possible combinations for different token pooling  
	and attention sharing mechanisms are constructed.  
	Based on the proposed compact set, the number of tokens in each layer and the choices of layers sharing attention can be treated as hyper-parameters that are learned from data automatically. 
	Experimental results show that the proposed scheme can achieve up to $6.6\%$ accuracy improvement in ImageNet classification compared with the DeiT in~\cite{touvron2021training}.
\end{abstract}

\section{Introduction}
\label{section:intro}

Deep neural network (DNN) has been the core of the recent popular artificial intelligence (AI) wave. It greatly enhances the state-of-the-art (SOTA) for many tasks in natural language processing (NLP) and computer vision (CV), which are two major application fields of AI. Among them, transformer~\cite{vaswani2017attention, devlin2019bert, NEURIPS2020_1325cdae, shi2021sparsebert} has found its almost ubiquitous applications in the field of NLP due to its great advantage of long-range capture and parallelism capability compared to the previous prevalent recurrent neural network (RNN). 

\begin{figure}[t]
	\centering
	\includegraphics[width=1\linewidth]{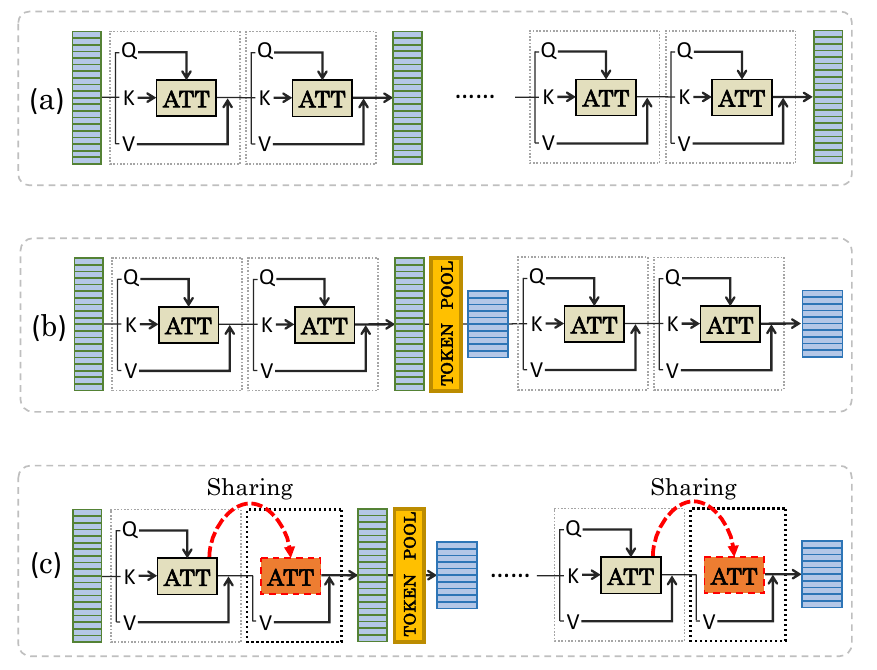}\\
	
	\caption{Token pooling and attention sharing. (a) The original ViT. (b) Token pooling  (TOKEN POOL in the figure) for ViT, where the number of tokens (input length) in the next layer can be smaller than the number of tokens in the current layer, and the token dimension per token can increase. (c) Attention map (ATT in the figure) sharing for ViT, where the attention at the previous transformer layer is copied to the current layer.}
	\label{fig:motivation}
\end{figure}

Recently, transformer is reevaluated by researchers in the field of CV and applied to solve some typical problems such as image generation~\cite{DBLP:journals/corr/abs-1802-05751} and image classification~\cite{dosovitskiy2020image, touvron2021training}. The flexible attention mechanism of the transformer in the whole pipeline intrigued the emergence of many following works such as  object detection~\cite{carion2020endtoend}, image segmentation~\cite{wang2020maxdeeplab}, and video instance segmentation~\cite{wang2020endtoend}. The inspiring achievements of the transformer in the above pioneering works are turning the transformer into a capable alternative to the prevalent convolutional neural network (CNN) in CV field.


In this paper, we find two redundancies in most vision transformers.
Our first observation is that the fixed number of tokens observed by the model produces the redundant features at the spatial level.
As shown in traditional CNNs~\cite{zeiler2013visualizing}, deep neural networks tend to encode low-level information in shallow layers and high-level information such as semantic features in deep layers. A fixed number of tokens may be an improper design for this purpose, since the fixed number of tokens may  not  be enough for low-level information encoding and too redundant for high-level information encoding. 

To better capture the semantic features as well as reducing the computation, it is a common practice to downsample feature resolution in CNN as the layers get deeper.
As a counterpart in a transformer, the number of tokens should have more flexibility to adjust itself based on different levels. To achieve this, we propose the attention pooling mechanism for adapting token numbers at different layers.

\begin{figure}[t]
	\centering
	\includegraphics[width=0.9\linewidth]{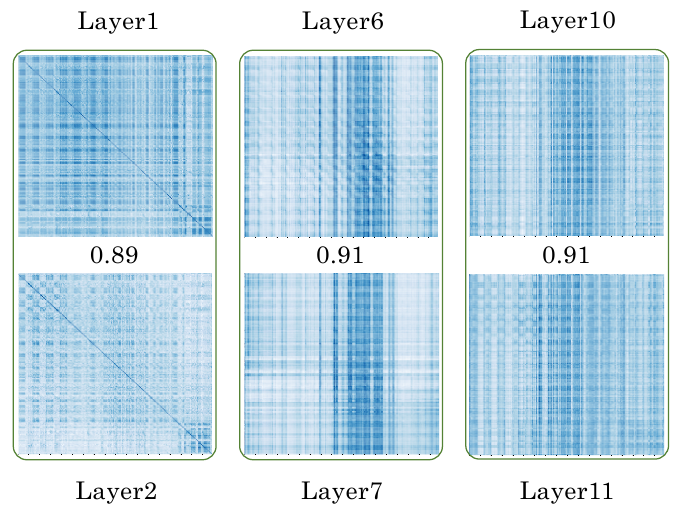}\\
	
	\caption{Attention maps from 6 different transformer layers. The attention maps of layers 1 and 2 have high normalized correlation, \ie 0.89. Similarly for layers 6, 7 and layers 10, 11 with the normalized correlation being 0.91.}
	\label{fig:attentionmap}
\end{figure}

The second form of redundancy resides in the similarity of the attention maps between adjacent layers. In ViT models, relations among all tokens are built in the transformer encoder according to the input feature embeddings. However, since feature embeddings between the neighboring layers change smoothly, 
the learned attention between neighboring layers may be similar as Fig.~\ref{fig:attentionmap} shows. 
The strong similarity between attention at adjacent layers motivates us to share the attention among adjacent layers. On the other hand, the attention takes 43\% computation and 33\% the number of parameters in the transformer layer. Sharing attention saves the computation required for obtaining attention.

Specifically, the number of tokens in each transformer layer and the setting for attention sharing will affect the performance of transformer. 
While it is intuitive for tuning token number and attention sharing to be effective, it is challenging to reach an optimal design for vision tasks. 
Instead of relying on heuristic human design for getting the optimal design, we can resort to a more principled way by leveraging the recent success of Automated Machine Learning (AutoML)~\cite{he2021automl}.
Specifically, we first construct a set of possible choices, called search space in AutoML, which consists of the possible design combinations for token pooling and attention sharing. 
For token pooling, we search its location in the network. Searching the locations of token pooling is equal to setting the optimal numbers of layers at different stages (we refer to the layers with the same number of tokens as a stage).


The choice of whether a layer shares the attention with its neighbour forms a discrete optimization problem. Our objective is to find the optimal configuration of attention sharing which minimizes the computational cost as well as maximizing the performance. In this work, we apply Neural Architecture Search to solve the configuration.

We treat these key factors in the network design as hyper-parameters and learn these parameters from data with the AutoML method. The two differences between the original ViT model and our proposed PSViT are shown as Fig.~\ref{fig:motivation}.


Our contributions can be summarized as follows:

\begin{itemize}
	\item {We present a PSViT scheme with token pooling and attention sharing to devise a better vision transformer that can effectively enhance the feature representation ability and control the computation allocation flexibly}
	
	
	\item {We propose a compact set of  design choices for token pooling and attention sharing, based on which the  optimal design choice is learned by leveraging off-the-shelf AutoML methods.}
	
	\item Comprehensive experimental results show that the proposed scheme has up to 6.6\% accuracy improvement in ImageNet classification compared with the DeiT in~\cite{touvron2021training}.
\end{itemize}

\begin{figure*}[t]
	\centering
	\includegraphics[width=1\linewidth]{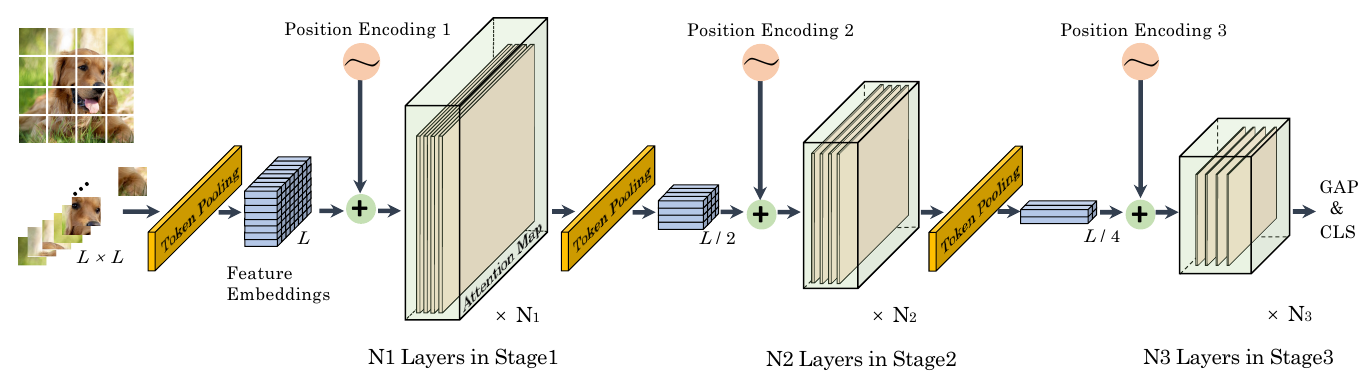}\\
	\caption{\by{Pipeline of our proposed PSViT model. There are three stages in the model, which are separated from the proposed token pooling layers. The token pooling layers help to reduce the spatial size (from $L$ to $L/4$ in the Figure) and increase the dimension of feature embedding. Each stage has several layers (including the layer with independent or sharing attention). The attention maps from different layers in the same stage have the same size. 
			We apply Global Average Pooling (GAP) to the last features and send the generated vector to a classifier (CLS) for the final classification. }}
	\label{fig:pipeline}
	
\end{figure*}


\section{Related Work}
\label{section:relatework}
\noindent\textbf{Transformers for Vision.}
Transformer~\cite{vaswani2017attention} was originally proposed to overcome the limited attention range and unsuitability of RNN in NLP. It then quickly gained popularity in the whole field of NLP. In the past few years, a new trend of applying transformer in the field of CV is witnessed. And transformer first found its application for low-level  vision problem such as image generation~\cite{DBLP:journals/corr/abs-1802-05751} and image super resolution~\cite{yang2020learning}. Parmar \etal~\cite{DBLP:journals/corr/abs-1802-05751}  advanced image transformer to generate images through constraining the self-attention mechanism only to local neighborhoods and achieved promising results.  Yang \etal~\cite{yang2020learning}  came up with a new texture transformer network for image super-resolution, where the low resolution and reference images are formulated as queries and keys in a transformer and obtained new SOTA for image super resolution.

Later, transformer was extended to high-level vision problems such as object detection~\cite{carion2020endtoend,zhu2020deformable}, image classification~\cite{dosovitskiy2020image,pmlr-v119-chen20s} and so on. To overcome the problem of some hand-designed components such as non-maximum suppression and anchor generation in the typical object detection, a complete end-to-end scheme (DETR)~\cite{carion2020endtoend} based on transformer was proposed with impressive results. However, DETR still suffers from   low convergence. As such,  Zhu \etal~\cite{zhu2020deformable} introduces Deformable DETR, in which attention modules only consider a small set of key sampling points, leading to much faster training speed and better performance especially for small objects. 
Dosovitskiy \etal~\cite{dosovitskiy2020image} explicitly defined the concept of Vision Transformer (ViT), and directly applied transformer to image classification with on-par-with or even better performance. Chen \etal~\cite{pmlr-v119-chen20s} utilized a sequence transformer to auto-regressively predict pixels with no need of 2D input structure to learn useful representations for images. The notable results reported by them 
demonstrate the attractive potential of transformers for unsupervised representation learning.  



Different from all the previous transformer based works in the field of CV, we concentrate on the token pooling and attention sharing that are not touched before. In addition, we apply the AutoML method to achieve this goal, which makes our work distinct from the manual design of these previous works.

\noindent\textbf{Automated Machine Learning (AutoML).}
Designing a better and efficient DNN is worthwhile for all the tasks in CV. Previously, most researchers and engineers in the field of DNN try to design better models based on heuristics and experiences. Such a process tends to be very tedious. 
Fortunately, in the past few years, a novel methodology of designing a better DNN model, that is, AutoML, arises quickly.  Currently, the AutoML methods for neural network architecture design can be divided into three categories, reinforcement learning (RL) based methods~\cite{DBLP:journals/corr/ZophL16,cai2017efficient, tian2020offpolicy}, neuro-evolutionary based  methods~\cite{elsken2019efficient,guo2020single, real2020automlzero} and differentiable methods~\cite{liu2018darts, DBLP:journals/corr/abs-1812-09926, wan2020fbnetv2}.

RL based methods~\cite{DBLP:journals/corr/ZophL16, cai2017efficient} and neuro-evolutionary methods~\cite{liu2018hierarchical, real2020automlzero} are good at searching discrete architectures, but generally requires huge computational cost in searching. To overcome the great computational burdens of RL and evolutionary methods during the search, differentiable search methods~\cite{liu2018darts, DBLP:journals/corr/abs-1812-09926, wan2020fbnetv2}  relaxed the discrete search space and applied SGD for searching. Then, it received a lot of attention due to its great advantage of searching efficiency~\cite{wan2020fbnetv2}, and it has been widely used in many computer vision problems such as segmentation~\cite{DBLP:journals/corr/abs-1901-02985,DBLP:journals/corr/abs-1912-10917} , object detection~\cite{DBLP:journals/corr/abs-1903-10979,ghiasi2019nasfpn,guo2020hitdetector},  model compression~\cite{dong2019tas, DBLP:journals/corr/abs-1812-00090, guo2020dmcp} and so on.

Inspired by the outstanding performance of AutoML for both CV and NLP, we also utilize AutoML for better ViT models with the aim of boosting its performance effectively. Specifically, the evolution based method of SPOS ~\cite{guo2020single} is taken as our hyper-parameters learning method due to its fast efficiency of the single path in the supernet searching and training. Nevertheless, AutoML has not been investigated for Transformer in image recognition and our search space of token pooling and attention sharing has not been studied in AutoML for CV.

\section{Method}
First, we give a brief introduction about transformer in Section~\ref{section:method-background}. Then, we present the two essential mechanisms of our PSViT, including token pooling and attention sharing, in Section~\ref{sub:tpst}. Finally, we introduce the enhanced PSViT with AutoML method in Section~\ref{section:automl}.


\subsection{Background about Transformer}
\label{section:method-background}
Transformer~\cite{vaswani2017attention} 
is built with stacked self-attention layers and feed-forward neural networks. 

\noindent\textbf{Self-attention Layer.} 
The self-attention layer utilizes the relevance or interaction of one token to other tokens. The input of self-attention layer is transformed into three different vectors, \ie  the query vector $q$, the key vector $k$ and the value vector $v$, and their corresponding packing form from different inputs are $Q$, $K$, $V$ respectively, we can calculate their attention among different inputs by the following equation:

\begin{equation}
	\vspace{-4pt}
	Att(Q, K, V)  = Softmax(\frac{Q\cdot(K^T)}{\sqrt{d_h}} )\cdot V,
	\label{eq:attention}
\end{equation}
where $d_h$ is the number of features per token and $\cdot$ denotes matrix multiplication.
The equation above means that the self-attention can be obtained by first computing the dot product of the query with all the keys, \ie $Q\cdot(K^T)$. Then, they are normalized with the softmax function to get attention scores. Finally, the attention scores are multiplied by the value vector. As such, the output of each token becomes the weighted sum of all the tokens in a sequence/image.
To boost the performance of vanilla self-attention, multi-head attention is further proposed by applying multiple attention functions to the input.


\noindent\textbf{Vision Transformer.} Dosovitskiy \etal~\cite{dosovitskiy2020image} proposed ViT which directly applied the above transformer from NLP to the image classification task. They divided an image into a sequence of flattening 2D tokens that are treated as the inputs to transformer. A trainable linear projection was adopted for token embeddings with the aim of yielding constant widths for different layers in a transformer. Then, another learnable embedding is further applied to the sequence of embedding tokens. In addition, 1D positional encoding is also attached to the token embeddings to maintain the positional information. 
When ViT is trained on large datasets, it can achieve a very impressive classification accuracy of 88.36\%. 

Building upon this work, Touvron \etal~\cite{touvron2021training} improved the ViT training speed by  introducing a novel teacher-student token distillation strategy specific to transformers (DeiT), yielding competitive results on ImageNet without external data as in ViT. 
We choose DeiT as our baseline model in this work for its great training efficiency.



\begin{table*}[!t]
	\centering
	\caption{Preliminary experiments on ImageNet about Token Dimension Settings for Each Stage. `Token Dimensions' denotes the channel number of tokens for the three stages. Similarly for `Token Numbers' and `Layer Numbers'.}
	\vspace{5pt}
	
	\begin{tabular}{|c|c|c|c|}
		\hline
		Model & DeiT-Tiny & Token Dimension1 & Token Dimension2 \\
		\hline
		w/o Pooling & no Pooling   &  2 Poolings &   2 Poolings\\
		\hline
		Token Dimensions  & {[}192, 192, 192{]} & {[}192, 192, 192{]} & {[}192, 256, 384{]} \\
		\hline
		Token Numbers  & {[}197, 197, 197{]} & {[}197, 99, 50{]} & {[}197, 99, 50{]}\\
		\hline
		Layer Numbers  & {[}4, 4, 4{]}  & {[}4, 8, 20{]} & {[}4, 4, 4{]} \\
		\hline
		FLOPS(G) & 1.3 & 1.3 & 1.3\\
		\hline
		Top1 Acc &  72.2\% & 75.0\% & 76.3 \% \\
		\hline
		
	\end{tabular}
	\label{tab:attpooling} 
\end{table*}

\subsection{Token Pooling and Sharing for Transformer}
\label{sub:tpst}
Based on the above review and problem analysis for the current ViT, we propose two mechanisms named token pooling and attention sharing to overcome the above three issues for the goal of a better ViT.

\subsubsection{Token Pooling}
\label{section:token-pooling}

The token pooling mechanism  adjusts the number of tokens for each stage as Fig.~\ref{fig:pipeline} shows. 
When the network depth increases, we decrease the number of tokens for removing spatial redundancy and increase the feature dimension for accommodating more high-level features that should be different from each other.
\by{There are two design options for the token pooling. The first one is following the network design in~\cite{dosovitskiy2020image}, treating the image patches as 1D tokens and utilizing the additional CLS token for the classification task. The other one is removing the CLS token and keeping the image patches in a 2D array, which is the same as the pooling strategy in most networks for computer vision tasks, such as ResNet~\cite{DBLP:journals/corr/HeZRS15}.}

For the first design, similar to~\cite{informer},  we achieve token pooling by convolution and maxpooling for the convenience of implementation. Different from~\cite{informer} that only decreases the number of tokens, our goal is to enhance the feature representation ability. Therefore, the feature dimension is not changed in~\cite{informer} but will be changed in ours. Specifically, we first utilize 1D convolution with a small kernel size to change the feature dimension (i.e., dimension of each token) and then decrease the number of tokens via 1D maxpooling. \by{We name our network with the above pooling strategy as PSViT-1D. In the second strategy, we adopt a 2D convolutional layer with stride 2 for token pooling, which is widely applied in many convolutional networks. Here, we name the network with the second pooling strategy as PSViT-2D.}



Simply adding a token pooling layer after each layer will decrease the model's representation ability expeditiously. Motivated by the principle of deep networks in CV, such as VGGNet, ResNet and MobileNet, we add a pooling layer after several layers and name the layers with the same token numbers as a stage.

\noindent\textbf{Investigation on Two Choices of Token dimension.}
To increase the representation ability of the ViT while maintaining its computational overhead in terms of FLOPS at the same time, we may have two choices to modify the network structure: increasing token dimension while decreasing the token number or keeping the tokens dimension and increasing the transformer layer number. Based on these two choices, we conducted some preliminary experiments \by{
	with the 1D pooling strategy} in Table~\ref{tab:attpooling}, which shows that increasing the feature dimensions while decreasing the number of tokens is a better choice.
Specifically, in the experiments, the baseline model fixes the feature size while the `Dimension1' scheme in Table~\ref{tab:attpooling} fixes the token dimensions and decreases the token numbers through the stages. To balance the total computation, several encoder layers are added. The `Dimension2' scheme in Table~\ref{tab:attpooling} increases the token dimensions while decreasing the token numbers. 
These two schemes design the architectures following the principle that the computation is uniformly distributed across the stages~\cite{Liang2020Computation}. 
We can find in Table~\ref{tab:attpooling}, these two token dimension designs perform much better than the DeiT-Tiny model which fixes the feature sizes through the whole network. The method decreasing the token number while increasing the feature dimension (`Dimension2') achieves the best performance.

\noindent\textbf{Analysis on the Design Choices.} In CV tasks, there are two characteristics.
\emph{First}, as discussed in Section~\ref{section:intro}, high-level features have redundancy at spatial level. Therefore, it is reasonable to reduce the redundancy at the spatial level by token pooling. This is supported by the results in Table~\ref{tab:attpooling}, where two token pooling design performs much better than the DeiT-Tiny model without token pooling. 
\emph{Second}, low-level features could be few but high-level features should be more.
Different layers in a deep network encode information of different levels. Low-level features, e.g. edges and textures, at shallow layers can be few and can be shared for representing high-level features. In contrast, high-level features, e.g. attributes or objects of different viewpoints, at deeper layers are more difficult to be shared. Therefore, most of the CNN designs, such as VGGNet~\cite{simonyan2015deep}, ResNet~\cite{DBLP:journals/corr/HeZRS15} and MobileNet~\cite{sandler2019mobilenetv2}, follow the rule that deeper layers have higher feature dimensions. This is also validated by the results in Table~\ref{tab:attpooling}, where increasing the feature dimension at the deeper layer in `Dimension2' performs better than fixing the feature dimension in `Dimension1'. 

\subsubsection{Attention Sharing}
\label{section:atten-sharing}
As shown in Fig.~\ref{fig:attentionmap}, the attention maps in continuous multi-attention layers are similar. As one of the reasons, identity mapping is used in the transformer, making a transformer layer function as the residual. In this case, the main features, though changed by the residual, will not drastically change. Therefore, the features in adjacent layers will be similar to each other, leading to the attention maps among features similar to each other. This similarity leads to redundancy.
Hence, it is beneficial to apply some sharing mechanisms so as to reduce redundancy and, at the same time, decrease the computation.

Specifically, we achieve the goal of attention sharing through reusing the attention calculation process between adjacent layers as shown in Fig.~\ref{fig:motivation}(c).
If a layer reuses the attention scores from its preceding layer, then we do not calculate the attention scores as well as the $Q$ and $K$ in Eqn.~(\ref{eq:attention}). The attention calculated at the preceding layer is used as the input attention of this layer and the whole process of attention calculation in Eqn.~(\ref{eq:attention}) is simplified to the dot product between $V$ and input attention scores.

Attention sharing can help to remove the redundancy of the attention map among adjacent transformer layers. On the other hand, some adjacent layers might have very different features and sharing their attention maps becomes not so effective. Taking this into consideration, we should provide the flexibility so that 
the whole ViT still has the choice of using the original multi-head attention module without sharing attention maps. As such, we take the sharing attention module as an optional choice to the original independent multi-head attention module in designing the overall transformer architecture.

\subsection{AutoML Enhanced Transformer}
\label{section:automl}

Recently, AutoML has greatly boosted the SOTA for many models in CV and NLP. In this section, we also turn to AutoML for the goal of a better ViT. 

\subsubsection{Search Space for PSViT}
\label{section:search-space}

As mentioned before, token pooling refers to the operation that decreases the number of tokens at the spatial level and increases the channel number of features per token, and such an operation may greatly affect the feature representation ability. Hence, we specifically consider its different forms in ViT.
The design choices of each stage mainly include three factors: the number of tokens $N_{t}$, the token dimension $N_{f}$, and the number of layers for each stage $N_{b}$.  To get an optimal choice for these three factors, we construct a search space that has multiple choices for them. Each element in the search space would lead to a new candidate network architecture for the transformer. 

In each layer, there are $S_t$ possible choices for the number of tokens, $S_f$ possible choices for the token dimension, and $S_s$ possible choices for using the attention maps for each map. For $L$ layers, the search space considering only the token pooling would contain $(S_t\cdot S_f \cdot S_s)^L$ candidate designs. For example, when $S_t= 4 $, $S_f= 4 $, $S_s= 4 $, and $L=36$, then there are about $1.1*10^{65}$ choices, about $9.6*10^{52}$ times the size of SPOS method~\cite{guo2020single}. In comparison, SPOS has 4 choices in each layer, while this search space has 64 in each layer. This search space is too large and has too many choices in each layer. It costs too much time when searching algorithms are used for obtaining the best architecture from this search space. Besides, too many choices in each layer may make the existing fast searching algorithm not effective, as found in~\cite{ci2020evolving}. Therefore, we need to reduce the search space.

Our designs of the search space and reasons are as follows:
\begin{itemize}
	
	
	\item{According to the analysis in Section~\ref{section:token-pooling}, we increase the token dimension but decrease the token numbers as the depth increases. This helps to reduce the search space by removing the network architectures that do not fit this rule.}

	\item Following the mature design of CNN, we constrain that multiple layers within a stage have the same number of tokens and the same token dimension. 
	To reduce the search space and computation required by the searching algorithm, we only use three stages of token pooling.
	
	\item{For the number of layers at each stage, we only consider the limited number of layers to further reduce the search space and computation. 
	}
	
	\item{We provide flexibility to apply attention sharing or not for different layers.}
	
\end{itemize}
%

\begin{figure}[!t]
	\centering
	\includegraphics[width=0.8\linewidth]{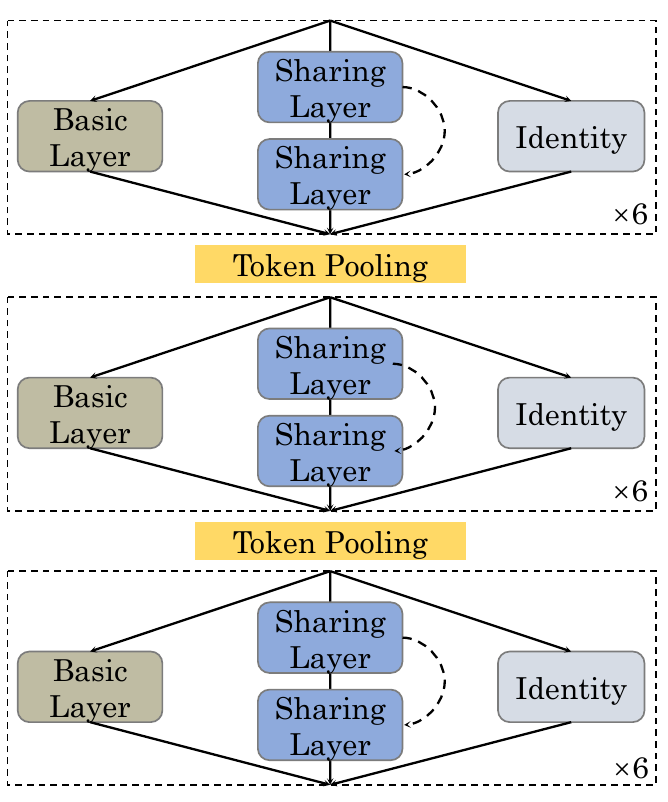}\\
	\caption{The architecture of our supernet. There are three stages, each stage containing 6 cells. A cell has three choices of paths, a basic transformer layer, sharing layers, and an identity layer. A candidate network has a single path for each cell.}
	\label{fig:supernet}
\end{figure}

\noindent\textbf{Our supernet design  for Representing the Search Space.}
If we consider the number of layers and the setting for attention map sharing in each stage as hyper-parameters for a network, then a natural choice for learning them is the RL or EA based approaches, which are very slow in searching. Therefore, we employ weight sharing based methods~\cite{cai2019proxylessnas,pmlr-v80-bender18a} as our searching algorithm, which requires defining a supernet to subsume all candidate architectures in the search space.
A possible design of supernet is shown in Fig. ~\ref{fig:supernet}. There are three stages in the supernet. A token pooling layer is placed between stages to change the token number and features dimension. Each stage has six cells, where a cell has three choices of paths, a basic transformer layer, two sharing layers with the attention map of the first transformer layer being copied to the second layer, and identity mapping. 
A candidate architecture could choose one path out of three choices for each cell independently. With the inclusion of identity mapping, the candidate network can have 0 (all cells choosing the identity layer) to 36 (all cells choosing the sharing layers) transformer layers.
Such supernet design provides more feasibility to the transformer architecture. For example, a candidate could only select basic layers for all layers, which is equivalent to the transformer enhanced by two token pooling layers. As another example, a candidate architecture could select the sharing layers path for all the first 6 cells at the first stage and select identity for the remaining cells, which has 12 transformer layers with every two layers sharing attention map but without token pooling.

\subsubsection{AutoML for the Searching}
With the search space and supernet defined in section \ref{section:search-space}, we find the optimal network architecture for the transformer using the SPOS~\cite{guo2020single}, which is briefly introduced below. 


\noindent\textbf{Training the supernet.}
The built supernet in section \ref{section:search-space} includes all the candidate networks. For each cell of the supernet, there are multiple choices. By activating only one choice in each cell, a candidate network can be constructed. 
During training, SPOS only selects one candidate network by uniform sampling for each training iteration and updates the parameters of the selected candidate network in the supernet. The candidate network will inherit the trained parameters from the supernet and keep updating these parameters. Once the supernet is trained, all the candidate networks can inherit the weights from it without being trained from scratch for searching.


\noindent\textbf{Searching from the trained supernet.}
After the supernet is trained, the best candidate network architecture can be obtained by further applying the evolutionary method to all the candidate networks in the supernet. More details could be found in~\cite{guo2020single}. 

\section{Experiments}

In this section, we elaborate on the  dataset, implementation details, and experimental results to evaluate the performance of the proposed PSViT. 

\subsection{Datasets and Implementation Details}
\label{sec:dataset}
\noindent \textbf{Experimental Dataset.} To validate the effectiveness of the proposed PSViT, we conduct experiments on image classification. We conduct all classification experiments on the ImageNet 1000-class image classification dataset, which is a widely used benchmark for validating network architecture design. This dataset has about 1 million training data. Part of the training data is used as the validation data. The training data without the validation data is used for training the supernet. The validation data is used by the AutoML for searching from the trained supernet.

\noindent \textbf{Implementation Details.} We choose the DeiT~\cite{touvron2021training} mentioned before as our baseline since it is the current SOTA for vision transformer in image classification. 
DeiT has two major different versions, that is, the Small one and the Tiny one, for efficient training, which are derived from the basic version of DeiT-B that has 12 transformer layers, a fixed token dimension of 768, 12 heads, and feature dimension of each head $d_h$ being 64. The differences between the Small one and the Tiny one are feature dimension and head number. More details about the other related details for DeiT can be found in~\cite{touvron2021training}.  
We follow the framework of DeiT~\cite{touvron2021training} to obtain the feature embeddings, which are sent to transformer encoders for further processing. 
We implemented our model based on PyTorch.

\noindent {\bf{Training settings. }} 
For the supernet training, mini-batch Nesterov SGD optimizer with a momentum of 0.9 is adopted.  
The initial learning rate is set to 0.2 and adjusted to 0 gradually by the cosine annealing learning rate decay. We train the network with a batch size of 1024 and L2 regularization of 1e-4 for 100 epochs. Besides, the label smoothing is applied with a 0.1 smooth ratio. For subnet searching, we use the same setting for the evolutionary algorithm as that in~\cite{guo2020single} under the FLOPS constraint. Specifically, the evolutionary algorithm samples $N_s = 1000$ subnets with population size 50, max iterations 20, and retaining top-10 models during the evolutionary search. 
After the optimal subnet is found, we retrain the subnet.
For retraining the subnet, we utilize the same training setting as that in~\cite{touvron2021training} due to its efficiency. 

\begin{table}[!t]
	\centering
	\caption{Comparison of ImageNet classification performance between different models. `Acc' denotes the Top-1 accuracy and `*' denotes the performance under our training setting, which is the same as that in DeiT~\cite{touvron2021training}.}
	\vspace{5pt}
	\centering
	\begin{tabular}{l||c|c}
		\hline
		Model  & {FLOPS(G)} & {Acc(\%)}   \\
		\hline
		\hline
		ResNet18  & {1.8}& {69.8}   \\
		ResNet18$^{*}$ & {1.8}& {68.5}   \\
		DeiT-Tiny  & {1.3}& {72.2}   \\
		PSViT-1D-Tiny  & {1.4}& {\textbf{77.4}}   \\
		PSViT-2D-Tiny  & {1.3}& {\textbf{78.8}}   \\
		\hline
		ResNet50 & {4.1}& {76.1}   \\
		ResNet50$^{*}$  & {4.1}& {78.5}   \\
		X50-32x4d  & {4.3}& {77.6}   \\
		X50-32x4d$^{*}$ & {4.3}& {79.1}   \\
		DeiT-Small  & {4.6}& {79.6}   \\
		PSViT-1D-Small  & {4.9}& {\textbf{80.7}}   \\
		PSViT-2D-Small  & {4.4}& {\textbf{81.6}}   \\
		\hline
		X101-64x4d  & {15.6}& {79.6}   \\
		X101-64x4d$^{*}$  & {15.6}& {81.5}   \\
		ViT-Base  & {17.6}& {77.9}   \\
		DeiT-Base  & {17.6}& {81.8}   \\
		PSViT-1D-Base  & {18.9}& {\textbf{82.6}}   \\
		PSViT-2D-Base  & {15.5}& {\textbf{82.9}}   \\
		\hline
	\end{tabular}
	\label{tab:ImageNet} 
\end{table}


\subsection{Classification Results}

For the classification task, we compare our two type searched models, PSViT-1D and PSViT-2D with two widely applied models, ResNet~\cite{DBLP:journals/corr/HeZRS15} and ResNext~\cite{xie2017aggregated}, and the SOTA pure transformer vision model DeiT~\cite{touvron2021training}.
More specifically, ResNet18 and ResNet 50 are shown in Table~\ref{tab:ImageNet} since they are provided in~\cite{touvron2021training} and they have similar FLOPs as DeiT-Tiny and DeiT-Small respectively, as pointed out in~\cite{touvron2021training}. Moreover, ResNetXt50-32x4d and ResNetXt101-64x4d that improved ResNet by aggregating the residual module are also included for comparison due to their outstanding performance in image classifications. For a fair comparison, the above corresponding models trained on our machine using our training settings (the same as DeiT) are further represented with a symbol of * in the table. 
The results for ViT-Base in Table~\ref{tab:ImageNet} is copied from~\cite{dosovitskiy2020image}. Note that using the training setting of DeiT for both ViT and DeiT will make ViT and DeiT having the same accuracy because the only difference between DeiT and ViT is the training setting.


Table~\ref{tab:ImageNet} shows the experimental results comparing our searched models with other methods under different computation budgets.   We can find that under similar FLOPS, our models achieve better classification accuracy.
For example, the accuracy of our searched 1D/2D model under 1.3 GFLOPS is more than 5\% / 6\% higher (we use absolute accuracy improvement instead of relative improvement in this paper) than DeiT-Tiny with similar FLOPS and more than 7\% / 8\% higher than the ResNet18, whose FLOPS is 1.4 times of ours. Such huge improvements in terms of better classification accuracy and fewer FLOPS make the proposed PSViT a more attractive and powerful competitor to the recent ViT and DeiT.

For ResNet50, ResNetXt, and DeiT-Small, our corresponding improved model obtained by the PSViT-1D also outperforms them respectively by a significant margin of 2.2\%,  1.6\%, and 1.1\%. The PSViT-2D is 0.9\% better than the PSViT-1D with fewer FLOPS. 
While for the large model comparisons among ResNetXt-101, DeiT-Base, and our base one, our respective classification gain is more than 1\% for the PSViT-2D-Base model. 

All the above three results for image classification show the effectiveness of the proposed PSViT scheme. We attribute the gain of our searched models to the two key parts advanced in this work: the proposed token pooling module that can more effectively capture different levels of image features, and the attention sharing mechanism that flexibly adapts the computational cost for different transformer layers, achieving a better trade-off between accuracy and model redundancy.






\begin{table*}[!h]
	\centering
	\caption{ImageNet top-1 classification accuracy for models with different mechanisms, \ie attention sharing, token pooling, and using AutoML or manual design for these two factors.}
	\vspace{5pt}
	\centering
	\begin{tabular}{cccccc}
		\hline
		{Attention} & {Token}& {AutoML}&  {FLOPS } & {acc-1D}  & {acc-2D}  \\
		{Sharing} & {Pooling}& {}&  {(G) } & {\%} & {\%}  \\
		\hline
		\hline
		& &  &  1.3&    72.2 & 72.2\\
		\hline
		\checkmark & &  &  1.3&   73.8 &  - \\
		\hline
		& \checkmark &  & 1.3&   76.3  & 76.7\\
		\hline
		\checkmark& \checkmark& \checkmark& 1.3&  77.4 & 78.8 \\
		\hline
	\end{tabular}
	\label{tab:mechanisms} 
\end{table*}

\begin{table*}[!t]
	\centering
	\caption{Token Number Setting based on PSViT-1D. Tiny/8 and Tiny/16 have differences in the input token numbers. To ensure Tiny/8 and Tiny/16 have similar FLOPs, token dimension and number of heads in the multi-head attention are adjusted correspondingly. Similarly for Small/8 and Small/16. `Token Dimensions' denotes numbers of features per token for the three stages, similarly for `Token numbers'. `Num. heads' denotes the number of heads in the multi-head attention for the three stages.}
	\vspace{5pt}
	\centering

	\begin{tabular}{|c|c|c|c|c|}
		\hline
		Model & Tiny/8  & Tiny/16 & Small/8   & Small/16  \\
		\hline
		\hline
		
		Token Dimensions  & [64, 144, 192] &  [192, 288, 384] & [144, 256, 384] & [288, 512, 768]\\
		\hline
		{Num. heads} & [1, 3, 3] & [3, 6, 6]& [3, 4, 6]  & [6, 8, 12] \\
		\hline
		Token Numbers  & [785, 393, 197] & [197, 99, 50] & [785, 393, 197] &   [197, 99, 50] \\
		\hline
		{Params(M)} & 3.8  & 15.6 & 13.9 & 53.8\\
		\hline
		FLOPS(G) & 1.5 & 1.3 & 4.9 & 4.0\\
		\hline
		Top1 Acc &  72.71\% & 77.40\% & 80.70 \% & 78.32 \%  \\
		\hline
	\end{tabular}
	
	%
	\label{tab:pooling} 
\end{table*}

\subsection{Ablation Study} 
To have a better understanding of the proposed PSViT, we carry out some ablation studies in this section, which are detailed as follows. All experimental results are conducted on ImageNet using the same training setting mentioned in Section~\ref{sec:dataset}.


\noindent \textbf{Effectiveness of Token Pooling, Attention Sharing, and AutoML.}
First of all, we evaluate our proposed mechanism through simple manually designed models as in Table~\ref{tab:mechanisms}. We first devise the ViT model with attention sharing between two adjacent layers. To balance the computation, we add additional layers at the end of the original model to reach similar FLOPS. The attention sharing improves the accuracy by 1.6\%.
Then, we evaluate the proposed token pooling with the setting of `Token Dimension2' described in Table~\ref{tab:attpooling}. Token pooling improves the performance of the model significantly, that is, from 72.2\% to 76.3\% for PSViT-1D and to 76.7\% for PSViT-2D. 

Moreover, we study the effectiveness of AutoML. We utilize SPOS method to search for an optimal architecture under the FLOPS constraint, considering both token pooling and attention sharing. The supernet for the SPOS is shown in Fig.~\ref{fig:supernet}. As shown in Table~\ref{tab:mechanisms}, there is  obvious performance improvement of 1.1\%/2.1\% for PSViT-1D/PSViT-2D, which demonstrates the efficacy of the utilized AutoML method of SPOS.

\noindent \textbf{Token Number Settings.}
As we mentioned before, the hyper-parameters about attention pooling have multiple choices. We evaluate the different settings as in Table~\ref{tab:pooling}.
Since we follow the principle of increasing token dimensions while decreasing token numbers, the only hyper-parameters should be the initial/final token numbers. We evaluate two types of token numbers based on PSViT-1D, corresponding to the way that input images are split. For a model with input token having size $16 \times 16$ (models with `/16'), the initial/final tokens number is 197/50 while the model with input token size being $8 \times 8$ (models with `/8') has 785/197 initial/final tokens. The model Tiny/16 is our searched model as mentioned and other models are obtained through scaling up (Small/16) or tuning the initial token numbers (Tiny/8, Small/16). 
For the Tiny models, the model with fewer token numbers performs much better. 
The Tiny model with more tokens has too many tokens with lower feature dimension in the final stage which leads to poor representation ability. Different from Tiny model, the Small model with more tokens achieves better performance. The reason is that the token dimensions are enough for the small model and the performance of the small model with few tokens drops due to the possible overfitting 
caused by the superfluous parameters.

\noindent \textbf{Attention Sharing Settings.}
We further study different attention sharing paradigms. And the experimental results based on 1D pooling are shown in Table ~\ref{tab:sharing}. In the table, `sharing 2' means every 2 adjacent layers share the same attention. `sharing 3' denotes every 3 adjacent layers share the same attention, \ie more layers sharing the same attention. 
We can clearly see that applying sharing can yield obvious improvements in classification accuracy compared to the vanilla ViT that does not share attention (`no sharing' in Table~\ref{tab:sharing}), with a promising margin of 1.6\%, and 0.9\% respectively for sharing in 2 layers and 3 layers. Sharing between 2 layers achieves better performance than sharing among 3 because features will change more when going through 3 layers and the attention may be different in  3 layers. For a better performance, we select sharing within 2 layers as our network design element in the supernet. 

\begin{figure*}[!h]
	\centering
	\includegraphics[width=1\linewidth]{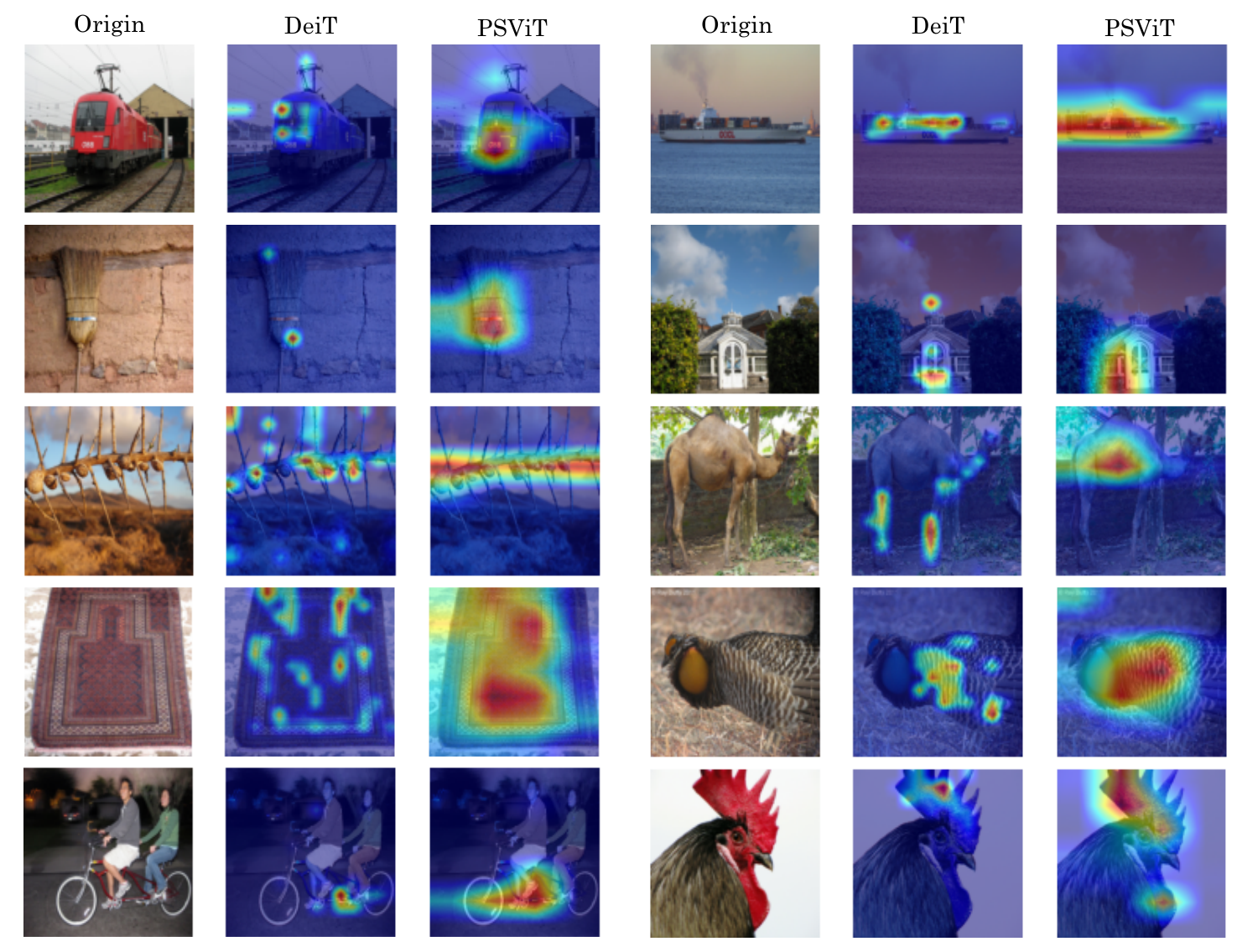}\\
	\caption{Visualization of features for DeiT and our PSViT. Images in the 1st and 4th columns are from ImageNet.
	}
	\label{fig:vis}
\end{figure*}

\begin{table}[!h]
	\centering
	\caption{ImageNet top-1 classification accuracy for  different sharing settings. `no sharing' denotes the network without sharing attention. `sharing 2' and `sharing 3' respectively denote every 2 and 3 adjacent transformer layers share the same attention map.}
	\vspace{5pt}
	\centering
	\begin{tabular}{c|cc}
		\hline
		model &  {FLOPS (G) } & {acc-1D (\%)}  \\
		\hline
		\hline
		no sharing & 1.3 &  72.2   \\
		sharing 2  & 1.3 &  73.8  \\
		sharing 3  & 1.3 &  73.1  \\
		\hline
	\end{tabular}
	\label{tab:sharing} 
\end{table}

\subsection{Visualization}
\noindent {\bf{Visualization.}} 
\by{In Fig.~\ref{fig:vis}, we show the visualization of learned features from DeiT (the 2nd and 5th columns) and our PSViT (the 3rd and 6th columns). We utilize the Grad-CAM in~\cite{Grad-cam} to show where is `important' for predictions. We can find that our PSViT can focus on a more complete region-of-interests than DeiT.} 
\bp{In addition, our model has larger feature dimensions which can capture more information than the baseline model, leading to possible better classification or recognition results.}

\begin{figure*}[!h]
	\centering
	\includegraphics[width=1\linewidth]{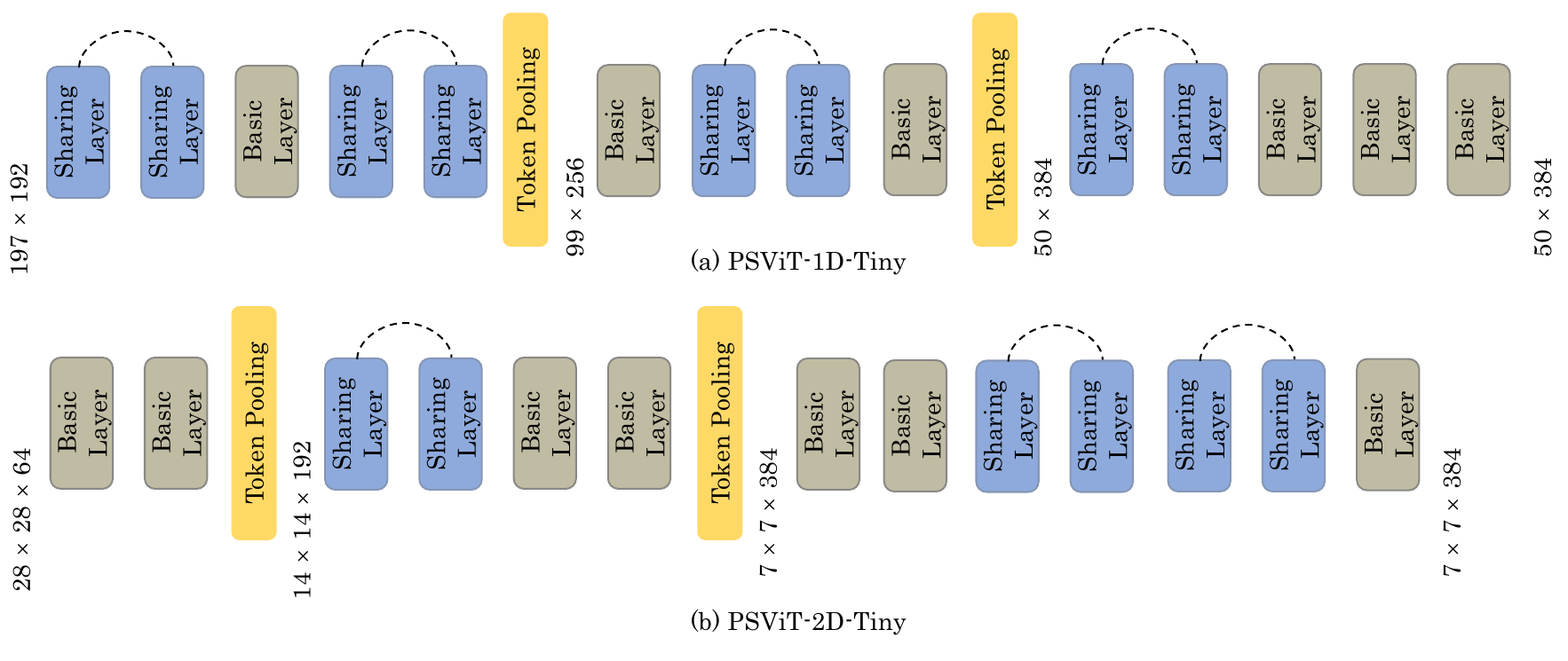}\\
	\caption{Searched architectures of PSViT-1D-Tiny and PSViT-2D-Tiny. \by{The feature size does not change in the same stage.}
	}
	\label{fig:arch}
\end{figure*}

\subsection{Searched Architectures}
\by{Fig.~\ref{fig:arch} shows the searched architectures of PSViT-1D and PSViT-2D. There are about 50\% layers sharing the attention, which shows that the proposed attention sharing mechanism can achieve a great trade-off between accuracy and computation costs. For the computation allocation among the three stages, both models allocate less computation in the first stage and more computation in the last stage. The features in the first stage have a larger spatial size. Applying self-attention to these features leads to a heavy computation burden. We think that is why the first stage tends to adopt attention sharing more times or has fewer total layers. It is a good strategy to reduce unnecessary computation.
	Differently, applying self-attention on deeper layers which have less tokens is better. The saved computation allows us to increase the feature dimension or layer number properly for a stronger feature representation.  
	For the last layer of the two models, there is no attention sharing. The last layer deserves an independent self-attention operation because its output is directly used for the final classification. More computation on the last layer will produce more accurate classification results.}

\begin{table*}[!h]
	\centering
	\caption{Performance comparison between different backbones on the object detection and instance segmentation tasks.}
	\centering
	\begin{tabular}{|c|c|c|c|c|c|c|c|c|c|c|c|c|}
		\hline
		\multirow{2}*{Backbone} & \multicolumn{6}{c|}{Object Detection } & \multicolumn{6}{c|}{Instance Segmentation} \\
		\cline{2-13}
		&  $AP$ & $AP50$ & $AP75$ & $AP_{s}$ & $AP_{m}$ & $AP_{l}$&  $AP$ & $AP50$ & $AP75$ & $AP_{s}$ & $AP_{m}$ & $AP_{l}$   \\
		\hline
		ResNet18 & 34.8 & 56.3 & 37.5 & 20.1 & 37.0 & 45.2 & 32.8 & 53.4 & 34.8& 17.4 & 25.1 & 44.7     \\
		\hline
		PSViT-2D-Tiny  & 40.8 & 64.7 & 44.0 & 25.3 & 43.8 & 53.9 & 37.7 & 60.1 & 39.9 & 21.2 & 40.6 & 52.8    \\
		\hline
		ResNet50  & 38.3& 60.4 & 41.4 & 23.3 & 41.6 & 48.9 & 35.5 & 57.2 & 37.8 & 19.6&38.6 & 47.7   \\
		\hline
		ResNet101 & 39.5 & 61.3 & 43.0 & 23.4& 42.7 & 51.1 & 36.5 & 58.2 & 39.1 & 19.6 & 39.7 & 49.4     \\
		\hline
	\end{tabular}
	\label{tab:det} 
\end{table*}

\subsection{Comparison between PSViT-1D and PSViT-2D}

\by{Our PSViT-1D and PSViT-2D have different searched architectures as shown in Fig.~\ref{fig:arch}. For PSViT-1D, we set the output token number as 50 (with the CLS token). Similarly, the spatial size of the output feature in PSViT-2D is $7 \times 7$. Both models have token pooling layers with a stride of 2, which means token number will reduce 50\% in PSViT-1D and 75\% for the PSViT-2D. In PSViT-1D, the token pooling is applied to 1D features. The token number is 197 for the first stage and 99 for the second stage. For PSViT-2D, the token number of the first stage and the second stage are $28 \times 28$ and $14 \times 14$, respectively. As the Table~\ref{tab:ImageNet} shows, our PSViT-2D models achieve better performance than PSViT-1D models. We think the main reason is that 2D token pooling can keep the structure information of 2D images while the 1D token pooling only considers the neighbouring tokens on the x-axis but ignores those on the y-axis. Besides, 2D features from PSViT-2D is more suitable for downstream tasks. }

\subsection{Object detection and instance segmentation results}
To evaluate the transferability of our models, we contrast our PSViT-2D model to ResNet on the downstream tasks, \ie object detection and instance segmentation. We apply the Mask-RCNN-FPN~\cite{mask-rcnn-kaiming} as our framework. We conduct experiments on MSCOCO 2017 dataset, which consists of 118K training images and 5K validation images. We report the performance on the validation set. 

For the network training,  we first initialize the backbone parameters with our pre-trained weights on ImageNet and adopt Xavier initialization on other layers in the neck and heads. 
Our models are trained with the batch size of 16 on 8 V100 GPUs and optimized by AdamW with the initial learning rate of 0.0001, 0.05 weight decay. 
We follow the standard common 1x training setting (12 epochs) to train the whole Mask R-CNN models with our searched models and ResNet as the backbone.
During the training stage, we follow the multi-scale training setting in~\cite{sun2021sparse}: resizing  the shorter side of input between 480 and 800 while the longer side is at most 1333. During testing, the shorter side of the input is fixed as 800 pixels.  

\by{The results are shown in Table~\ref{tab:det}. Our PSViT-2D-Tiny achieves 40.8\% mAP on the object detection task and 37.7\% mAP on the instance segmentation task, which is 6 and 4.9 points higher than that of ResNet18, respectively. Our PSViT-2D-Tiny even achieves better performance than ResNet101. The much better performance than ResNets on both object detection and instance segmentation shows the great  transferability of our models.}





\section{Conclusions}
In this paper, we have introduced a new PSViT by making use of token pooling and attention sharing. To effectively enhance the performance of current ViT,  token pooling and attention  sharing  are first designed. Then, the search space for a PSViT stage is designed. Based on the search space, we further apply an efficient AutoML method, SPOS, to obtain the optimal module design. Extensive comparison experiments on the ImageNet image classification dataset show the effectiveness of the proposed PSViT since it can increase the classification accuracy a lot. The great performance on object detection and instance segmentation \bp{also validates} the transferability of our models.

{\small
\bibliographystyle{ieee_fullname}
\bibliography{egbib}
}

\end{document}